\documentclass[submission,copyright,creativecommons]{eptcs}
\providecommand{\event}{FMAS 2021} 
\usepackage{breakurl}             
\usepackage{underscore}           
\usepackage{soul}
\usepackage{graphicx}
\usepackage{booktabs}
\usepackage{multirow}
\usepackage{amssymb}
\usepackage{color}
\usepackage{pifont}
\usepackage{tabularx}
\usepackage{graphics}
\usepackage{makecell}
\usepackage{balance}
\usepackage{hyperref}
\usepackage{comment}
\usepackage{adjustbox}
\usepackage{filecontents}

\usepackage{siunitx}

\usepackage{accsupp}
\usepackage{xcolor}
\usepackage{showexpl}

\usepackage{array}
\usepackage{caption}
\usepackage{subcaption}
\newcommand{\PreserveBackslash}[1]{\let\temp=\\#1\let\\=\temp}
\newcolumntype{C}[1]{>{\PreserveBackslash\centering}p{#1}}
\newcolumntype{R}[1]{>{\PreserveBackslash\raggedleft}p{#1}}
\newcolumntype{L}[1]{>{\PreserveBackslash\raggedright}p{#1}}
\usepackage{listings, xcolor}
\usepackage{adjustbox}

\newcommand{\XComment}[1]{}
\usepackage{enumerate}
\usepackage[shortlabels]{enumitem}

\usepackage{xspace}

\usepackage{pgfplots}
\pgfplotsset{width=7cm,compat=1.8}

\usepackage{tcolorbox}
\usepackage{verbatim}
\usepackage{comment}

\title{QuantifyML: {H}ow Good is my Machine Learning Model?}
\author{Muhammad Usman
\institute{University of Texas at Austin, USA}
\email{muhammadusman@utexas.edu}
\and
Divya Gopinath
\institute{KBR Inc., CMU, Nasa Ames}
\email{divya.gopinath@nasa.gov}
\and
 Corina S. P\u{a}s\u{a}reanu
\institute{KBR Inc., CMU, Nasa Ames}
\email{corina.s.pasareanu@nasa.gov}
}
\def\titlerunning{QuantifyML: {H}ow Good is my Machine Learning Model?}
\def\authorrunning{Muhammad Usman, Divya Gopinath and Corina S. P\u{a}s\u{a}reanu}
\begin{document}
\maketitle

\begin{abstract}

The efficacy of machine learning models is typically determined by computing their accuracy on test data sets. However, this may often be misleading, since the test data may not be representative of the problem that is being studied.  
With \emph{QuantifyML} we aim to \emph{precisely} quantify the extent to which machine learning models have learned and generalized from the given data. Given a trained model, \emph{QuantifyML} translates it into a C program and feeds it to the CBMC model checker to produce a formula in Conjunctive Normal Form (CNF). The formula is analyzed with off-the-shelf model counters to obtain precise counts with respect to different model behavior. 
\emph{QuantifyML} enables i) evaluating  learnability by comparing the counts for the outputs to ground truth, expressed as logical predicates, ii) comparing the performance of models  built with different machine learning algorithms (decision-trees vs. neural networks), and iii) quantifying the safety and robustness of models. 
\end{abstract}

\section{Introduction}

Recent years have seen a surge in the use of machine learning algorithms in a variety of applications to analyze and learn from large amounts of data. For instance, decision-trees are a popular class of supervised learning that can learn easily-interpretable rules from data. They have found success in areas such as medical diagnosis and credit scoring~\cite{kuo2001data,bastos2007credit}. Deep Neural Networks (DNN) also have gained popularity in diverse fields such as banking, health-care,  image and speech recognition, as well as perception in self-driving cars~\cite{6296526,NIPS2012c399862d}. 

Such machine learning models are typically evaluated by computing their {\em accuracy} on held-out test data sets, to determine how well the model learned and generalized from the training data. However, this is an imperfect measure, as they may not cover well the desired input space. Furthermore, it is often not clear which learning algorithm or trained model is better suited for a particular problem (e.g., neural networks vs. decision trees), and simply comparing the accuracy of different models may lead to misleading results. It may also be the case that well-trained models may be vulnerable to {\em adversarial attacks} \cite{SzegedyZSBEGF13,PapernotMJFCS16,huang2020survey} or they may violate desired {\em safety} properties \cite{9081758}. It is unclear how to quantify the {\em extent} to which these vulnerabilities affect the performance of a model, as evaluating the model on the available test or adversarial data sets may again give imprecise results.

\begin{figure}[t]
\includegraphics[width=\linewidth]{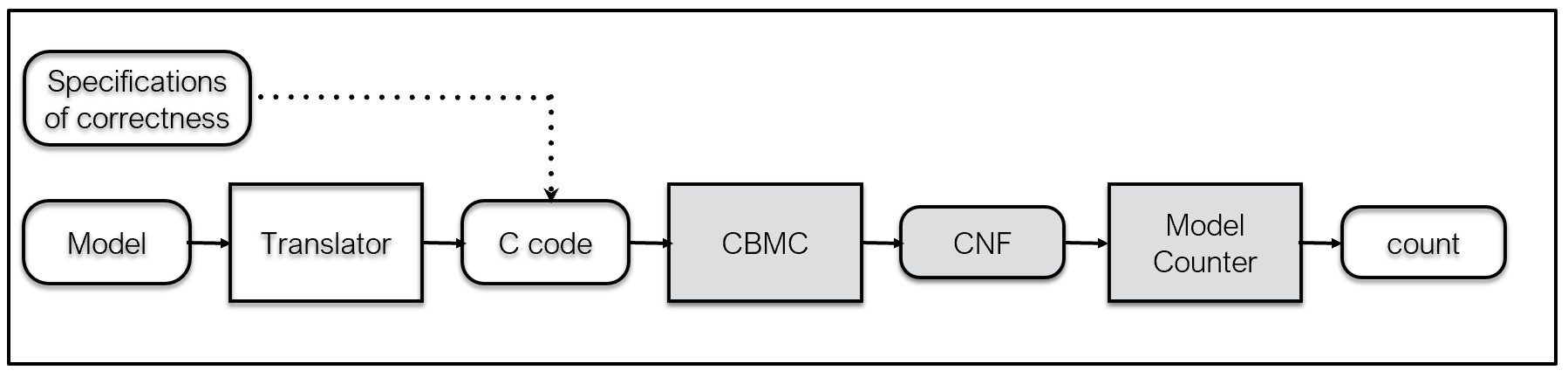}
\caption{\emph{QuantifyML} Framework \label{fig:framework}}
\end{figure}

We present \textbf{QuantifyML}, an analysis tool that aims to {\em precisely quantify} the learnability, safety and robustness of machine learning models. In this tool, a given trained model is translated into a C program, enabling the application of the CBMC tool \cite{kroening2014cbmc} to obtain a formula in Conjunctive Normal Form (CNF), which in turn can be analyzed with approximate and exact model counters \cite{gomes2008model,4268160,10.1007/3-540-46002-019} to obtain precise counts of the inputs that lead to different outputs.  Figure~\ref{fig:framework} gives a high-level description of \emph{QuantifyML}. 
We demonstrate \emph{QuantifyML} in the context of decision trees and neural networks for the problems of learning relational properties of graphs, image classification and aircraft collision avoidance. 

We derive inspiration from a recent paper \cite{10.1145/3385412.3386015} which presents \emph{Model Counting meets Machine Learning (MCML)} to evaluate the learnability of binary decision trees. With \emph{QuantifyML} we generalize MCML by providing a more general tool that can handle more realistic multi-class problems, such as decision trees with non-binary inputs and with more than two output decisions, and also neural networks. Other learning algorithms can be accommodated provided that the learned models are translated into C programs. \emph{QuantifyML}'s applications extend beyond MCML and include: (i) comparison of the performance of different models, built with different learning algorithms, (ii) quantification of robustness in image classifiers, and (iii) quantification of safety of neural network models.

\section{Background}
\label{sec:background}
 
\textbf{Decision Trees:}
Decision tree learning \cite{safavian1991survey} is a supervised learning technique for
extracting rules that act as classifiers. 
Given a set of data labeled to respective classes, decision tree learning aims to discover rules in terms of the attributes of the data to discriminate one label from the other. It builds a tree such that each path of
the tree encodes a rule as a conjunction of predicates on the data attributes. Each rule attempts to cluster or group inputs
that belong to a certain label. 


\noindent\textbf{Neural Networks:}
Neural networks~\cite{Goodfellow-et-al-2016} are machine learning algorithms that can be trained to perform different tasks such as classification and regression.
Neural networks consist of multiple layers, starting from the \textit{input} layer, followed by one or more \textit{hidden} layers (such as convolutional, dense, activation, and pooling), and a final \textit{decision} layer.
Each layer consists of a number of computational units, called \textit{neurons}. Each neuron applies an activation function on a weighted sum of its inputs;
$N(X)=\sigma(\sum_i w_i \cdot N_i(X) + b)$
where $N_i$ denotes the value of the $i^{th}$ neuron in the previous layer of the network and the coefficients $w_i$ and the constant $b$ are referred to as \emph{weights} and \emph{bias}, respectively; $\sigma$ represents the activation function.
The final decision layer (also known as \textit{logits}) typically uses a specialized function (e.g., max or \textit{softmax}) to determine the decision or the output of the network. 

\noindent\textbf{Bounded Model Checking for C programs:}
Bounded model checking~\cite{BMC} is a popular technique for verifying safety properties of software systems. Given a bound on the input domain and a bound on the length of executions, a boolean formula is generated that is satisfiable if there exists an error trace or counter-example to the given property. The formula is checked using off-the-shelf decision procedures.  CBMC~\cite{CBMC} is a tool that performs analysis of programs written in a high-level language such as C, C++ or Java by applying bounded model checking. The program is first converted into a control flow graph (CFG) representation and formulas are built for the paths in the CFG leading to assertions. The model checking problem is reduced to determining the validity of a set of bit-vector equations, which are then flatted out to conjunctive normal form (CNF) and checked for satisfiability. In this work, we leverage CBMC to build the CNF formulas corresponding to the paths in the C program representation of a machine learning model. We then pass on the formulas corresponding to the respective output classes to a model counting tool in order to quantify the number of solutions.



\noindent\textbf{Projected model counting:} Many tools, including CBMC, translate a boolean formula to CNF by introducing auxiliary variables. These variables do not affect the satisfiability of the boolean formula but do affect the model counts. In such scenarios, projected model counting~\cite{Aziz2015SATPM} needs to be used. Consider the set $M$ consisting of all variables in a boolean formula and $N$ be a subset of variables in the formula. The solutions in which the value of at least one variable in $N$ is different is considered a unique solution in the \emph{projected} model counting problem. The variables in $N$ are known as primary variables and the rest of the variables are known as auxiliary variables. Please refer to~\cite{usman2020testmc} for a detailed discussion on model counting, projected model counting and model counters. 
In our work we use projected model counting, where the inputs to the model are considered as primary variables. We used two state-of-the-art model counters i.e., projMC~\cite{projmc} and ApproxMC~\cite{10.1007/978-3-642-40627-0-18}.

\noindent\textbf{MCML:}
\emph{MCML} \cite{10.1145/3385412.3386015} uses model counting to perform a quantitative assessment of the performance of  decision-tree classifier models. The ground truth ($\phi$) is translated by the Alloy analyzer with
respect to bound $b$ into a CNF formula
$cnf_{\phi}$. 
It then translates the relevant parts of decision tree with respect to the
desired metrics (True Positives, False Positives, False Negatives, True Negatives) into a CNF formula $cnf_{d}$. It then combines these two formulas to create the CNF formula $cnf_{\phi,d}$ which is an input to
the model counter that outputs the number of solutions that satisfy the formula. This count quantifies the true performance of the decision tree. \emph{MCML} is limited to binary decision trees and has been used on decision-tree models when used to learn relational properties of graphs. \emph{QuantifyML} goes beyond \emph{MCML} as it enables quantification of the performance of more general machine learning models, that may have non-binary inputs and multi-class outputs. Our evaluation presents applications such as robustness analysis of decision-tree models on an image-classification problem (MNIST), comparison of neural network and decision-tree models for learning relational properties of graphs, and evaluation of safety for collision avoidance, which cannot be achieved with the \emph{MCML} tool.
\section{Approach}
\label{sec:approach}


\noindent{\bf Quantifying the learnability of machine learning models:}
\emph{QuantifyML} can be used to quantify the learnability of models, provided that a predicate is given which describes the ground-truth output for any input and finite bounds on the input space.  Consider a model classifying a given input into one of $L$ labels. For each output label $l$, two predicate functions are generated; $\phi_{l}(x)$ which returns 1 if the output of the model is $l$ for a given input $x$ and returns 0 otherwise, and $\psi_{l}(x)$ which returns 1 if the ground-truth for the given input $x$ is $l$ and returns 0 otherwise. These predicates are used to encode the following metrics for each label $l$; \textbf{True Positives (TP):}  $\; MC(\:CNF(\:\psi_{l}(x) \, \wedge  \, \phi_{l}(x)\: ), N)$, \textbf{False Positives (FP):}  $\; MC(\: CNF(\:\neg\psi_{l}(x) \, \wedge  \, \phi_{l}(x)\: ), N)$, \textbf{True Negatives (TN):}  $\; MC(\: CNF(\:\neg\psi_{l}(x) \, \wedge \, \neg\phi_{l}(x)\:), N)$, and \textbf{False Negatives (FN):}  $\; MC(\: CNF(\:\psi_{l}(x) \, \wedge \, \neg\phi_{l}(x)\:), N)$. $N$ is the scope or bound on the input domain,  $CNF$ represents a function that translates C program to formulas in the CNF form, and $MC$ represents a function that uses the projected model counter to return the number of solutions projected to the input variables. \emph{QuantifyML} then uses these counts to assess the quality of the model using standard measures such as \emph{Accuracy}, \emph{Precision}, \emph{Recall} and \emph{F1-score} for the model. \em Accuracy=$\frac{TP + TN}{TP+FP+TN+FN}$, \em Precision=$\frac{TP}{TP+FP}$, \em Recall= $\frac{TP}{TP+FN}$ and \em F1-score= $\frac{2*Precision*Recall}{Precision+Recall}$.

\noindent{\bf Quantifying the safety of machine learning models:}
\label{subsec:quantifysafety}
 \emph{QuantifyML} can also be used to quantify the extent to which input-output safety properties are satisfied for a model.
 Assume a property $p$ of the form ($Pre => Post$) where $Pre$ is a condition on the input variables and $Post$ is a condition on the output of the model, such as a classifier producing a certain label. We can use \emph{QuantifyML} to obtain the following counts: i) \emph{QuantifyML$_S$} denoting the portion of the inputs for which the model satisfies the given property, and ii) \emph{QuantifyML$_N$} denoting the portion of the inputs for which the model violates the property. These counts are then used to obtain an accuracy metric; \emph{QuantifyML$_{Acc}$}=$\frac{\emph{QuantifyML$_S$}}{\emph{QuantifyML$_N$}+\emph{QuantifyML$_S$}}$, which is a measure of the extent to which the network satisfies the property. 

\noindent{\bf Quantifying Local Robustness:}
The challenge with the analysis of more realistic models is that we typically do not have the ground truth. Image classification is such a problem, where it is not feasible to define a specification that can automatically generate the ground-truth label for any arbitrary image. However, images that are similar to, or are in close proximity (in terms of distance in the input space) to an image with a known label can be expected to have the same label. This property is called {\em robustness} in the literature. Current techniques \cite{gopinath2017deepsafe,10.1007/978-3-030-47358-71,CohenRK19} typically search for the existence of an adversarial input ($x'$) within an $\epsilon$ ball surrounding a labelled input ($x$); e.g., $|| x- x'||_{\infty} \leq \epsilon$ (here the distance is in terms of the $L_{\infty}$ metric) such that the output of the model on $x$ and $x'$ is different. When no such input exists, the model is declared robust, however, in the presence of an adversarial input there is no further information available.  \emph{QuantifyML}  can be used to {\em quantify} robustness of machine learning models, where instead of using a predicate encoding the ground truth, we encode the local robustness requirement that the model should give the same output within the region defined by $|| x- x'||_{\infty} \leq \epsilon$.
In order to quantify local robustness around a concrete n-dimensional input $x=(x_0,x_1, .. x_n)$, we first define an input region $R_\epsilon$ by constraining the inputs across each dimension to be within $[x_i-\epsilon,x_i+\epsilon]$ in the translated C program. We then define \emph{Robustness}$_{\epsilon}$ as 
$\frac{MC( CNF(\phi_{l}(x)), R_\epsilon)}{|R_\epsilon|}$,
where $\phi_{l}(x)$ is defined as before as a predicate which returns 1 if the output of the model is $l$ and 0 otherwise, $R_\epsilon$ defines the scope for the check, and $|R_\epsilon|$ quantifies its size. Intuitively, \emph{Robustness$_{\epsilon}$} quantifies the portion of the input on which the model is robust, within the small region described by $R_\epsilon$.

Please check longer version of this paper~\cite{PROJGITHUB} for more details on the approach. The tool currently supports decision trees trained using Scikit-Learn~\cite{pedregosa2011scikit} and neural networks trained in Keras~\cite{ketkar2017introduction}. 

\section{Evaluation}
\label{sec:evaluation}

We present experiments we have performed to evaluate the benefits of using \emph{QuantifyML} in the applications of quantifying learnability, safety and robustness of machine learning models. 

\noindent{\bf Quantifying the learnability of machine learning models:} 
This study aims to assess \emph{QuantifyML} in quantifying the true performance of models and enabling one to compare different models, and different learning algorithms, for a given problem. 

We evaluated the performance of trained models against ground truth predicates on the problem of learning relational properties of graphs. We considered 11 relational properties of graphs including Antisymmetric, Connex, Equivalence, Irreflexive, NonStrictOrder, PartialOrder, PreOrder, Reflexive, StrictOrder, TotalOrder and Transitive (refer \cite{PROJGITHUB}). We used the Alloy tool~\cite{Alloy} to create datasets containing positive and negatives solutions for each of these properties. Each input in the dataset corresponds to a graph with a finite number of nodes and is represented as an adjacency matrix. Each input has a corresponding binary label (1 if the graph satisfies the respective property, 0 otherwise). Please refer \cite{PROJGITHUB} for more details on the setup. The problem of learning relational properties of graphs albeit seems fairly simple with binary decisions and binary input features, it is not immediately apparent which learning algorithm would work best to learn a suitable classifier. We applied two different learning algorithms, decision-trees and neural networks, to learn classification models for the same set of properties using the same dataset for training. We were unable to apply \emph{QuantifyML} to analyze neural network models with greater than 16 features due to the limitation in the scalability of the model counters. Therefore we restricted the size of the graphs to have 4 nodes. 

\begin{table}[!ht]
\centering
\caption{Quantifying the learnability of Decision Trees on graph (4-node) properties with \emph{projMC}. \emph{Diff} shows the difference between Statistical (\emph{Stat}) and \emph{QuantifyML} (\emph{QML}) metrics.}
\vspace{-3mm}
\label{tab:4.2scores-DT}
 \begin{adjustbox}{max width=\textwidth}
\begin{tabular}{|l||c|c|c||c|c|c||c|c|c||c|c|c|}
	\hline

\multirow{2}{*}{\emph{Property}} & \multicolumn{3}{c||}{\emph{Accuracy}} & \multicolumn{3}{c||}{\emph{Precision}} & \multicolumn{3}{c||}{\emph{Recall}}&  \multicolumn{3}{c|}{\emph{F1-score}}\\ 
\cline{2-13}

& \emph{Stat} & \emph{QML} & \emph{Diff} &  \emph{Stat}  & \emph{QML} & \emph{Diff} & \emph{Stat} & \emph{QML}& \emph{Diff} & \emph{Stat}& \emph{QML} & \emph{Diff}\\ \hline

Antisymmetric&1.0000&1.0000&0.0000&1.0000&1.0000&0.0000&1.0000&1.0000&0.0000&1.0000&1.0000&0.0000\\
Connex&0.9932&0.8179&-0.1752&0.9865&0.4219&-0.5646&1.0000&0.0625&-0.9375&0.9932&0.1089&-0.8843\\
Irreflexive&1.0000&1.0000&0.0000&1.0000&1.0000&0.0000&1.0000&1.0000&0.0000&1.0000&1.0000&0.0000\\
NonStrictOrder&1.0000&0.9721&-0.0279&1.0000&0.1069&-0.8931&1.0000&1.0000&0.0000&1.0000&0.1932&-0.8068\\
PartialOrder&0.9957&0.9919&-0.0038&0.9916&0.8690&-0.1226&1.0000&1.0000&0.0000&0.9958&0.9299&-0.0659\\
PreOrder&1.0000&0.9693&-0.0307&1.0000&0.1499&-0.8501&1.0000&1.0000&0.0000&1.0000&0.2607&-0.7393\\
Reflexive&1.0000&1.0000&0.0000&1.0000&1.0000&0.0000&1.0000&1.0000&0.0000&1.0000&1.0000&0.0000\\
StrictOrder&0.9545&0.9721&0.0175&0.9200&0.1069&-0.8131&1.0000&1.0000&0.0000&0.9583&0.1932&-0.7651\\
Transitive&0.9850&0.9799&-0.0051&0.9810&0.7524&-0.2285&0.9904&0.9990&0.0086&0.9856&0.8583&-0.1273\\ \hline
\end{tabular}
\end{adjustbox}

\end{table}

\begin{table}[!ht]
\centering

\caption{Quantifying the learnability of Neural Networks on graph (4-node) properties with \emph{projMC}. \emph{Diff} shows the difference between Statistical (\emph{Stat}) and \emph{QuantifyML} (\emph{QML}) metrics.}
\vspace{-3mm}
\label{tab:4.2scores-NN}
 \begin{adjustbox}{max width=\textwidth}
\begin{tabular}{|l||c|c|c||c|c|c||c|c|c||c|c|c|}
	\hline

\multirow{2}{*}{\emph{Property}} & \multicolumn{3}{c||}{\emph{Accuracy}} & \multicolumn{3}{c||}{\emph{Precision}} & \multicolumn{3}{c||}{\emph{Recall}}&  \multicolumn{3}{c|}{\emph{F1-score}}\\ 
\cline{2-13}

& \emph{Stat} & \emph{QML} & \emph{Diff} &  \emph{Stat}  & \emph{QML} & \emph{Diff} & \emph{Stat} & \emph{QML}& \emph{Diff} & \emph{Stat}& \emph{QML} & \emph{Diff}\\ \hline

\emph{Antisymmetric}&0.8058&0.7614&-0.0445&0.7520&0.4211&-0.3309&0.9095&0.9093&-0.0002&0.8233&0.5756&-0.2476\\
\emph{Connex}&0.9658&0.7866&-0.1791&0.9359&0.2326&-0.7033&1.0000&0.0865&-0.9135&0.9669&0.1261&-0.8408\\
\emph{Irreflexive}&1.0000&1.0000&0.0000&1.0000&1.0000&0.0000&1.0000&1.0000&0.0000&1.0000&1.0000&0.0000\\
\emph{NonStrictOrder}&0.9773&0.9054&-0.0719&0.9583&0.0338&-0.9245&1.0000&0.9909&-0.0091&0.9787&0.0654&-0.9133\\
\emph{PartialOrder}&0.7803&0.8303&0.0500&0.8367&0.2002&-0.6364&0.7051&0.7260&0.0210&0.7652&0.3139&-0.4514\\
\emph{PreOrder}&0.9577&0.8825&-0.0753&0.9302&0.0433&-0.8870&1.0000&0.9803&-0.0197&0.9639&0.0829&-0.8810\\
\emph{Reflexive}&1.0000&1.0000&0.0000&1.0000&1.0000&0.0000&1.0000&1.0000&0.0000&1.0000&1.0000&0.0000\\
\emph{StrictOrder}&0.9545&0.9409&-0.0136&0.9200&0.0535&-0.8665&1.0000&1.0000&0.0000&0.9583&0.1016&-0.8567\\
\emph{Transitive}&0.7722&0.7903&0.0181&0.8063&0.1864&-0.6198&0.7404&0.7258&-0.0145&0.7719&0.2967&-0.4753\\ \hline 
\end{tabular}
\end{adjustbox}
\vspace{-5mm}
\end{table}

Tables~\ref{tab:4.2scores-DT} and~\ref{tab:4.2scores-NN} presents the results. 
We can observe the benefit of  \emph{QuantifyML} over pure statistical results ($Stat$) for both decision-tree and neural-network models. The decision-tree models for the Antisymmetric, Irreflexive, Reflexive, NonStrictOrder and PreOrder properties have accuracy and F1-scores of 100\%. However, the counts computed by \emph{QuantifyML} highlight that for the NonStrictOrder and PreOrder properties, the models in fact have less than 100\% accuracy and more importantly have poor precision indicating large number of false positives. The decision trees for StrictOrder seem to have the lowest accuracy and F1-score when calculated statistically. However, the \emph{QuantifyML} scores indicate that this is mis-leading and  the decision-tree for the Connex property has the lowest accuracy and F1-score. For the neural networks, in all cases except Irreflexive and Reflexive properties, the accuracies calculated using \emph{QuantifyML} highlight that the true performance is mostly worse and in some cases better (PartialOrder, Transitive) than the respective statistical accuracy metric values. The statistical results give a false impression of good generalizability of the respective models, while in truth the F1-scores are less than 50\% for most of the properties (refer \emph{F1-score} column in table~\ref{tab:4.2scores-NN}).

The models for the Irreflexive and Reflexive properties have 100\% accuracy and F1-score.
These are very simple graph properties. However, decision-tree models have the ability to learn more complex properties such as Antisymmetric and StrictOrder as well. Overall the decision-tree models seem to have better accuracy and generalizability than the respective neural network models.  Note, that while such a comparison can be done using the statistical metrics, their lack of precision may lead to wrong interpretations. For instance, for the StrictOrder property, the statistical accuracy, precision, recall and F1-scores are exactly the same for the neural network and decision-tree models, however, the corresponding \emph{QuantifyML} metrics highlight that for this problem, the decision-tree model is in fact better than neural network in terms of the true performance.

\begin{table}[!ht]
\centering
\caption{Quantifying robustness for the MNIST model.}
\vspace{-3mm}
\label{tab:4.3scores}
\resizebox{\textwidth}{!}{
\begin{tabular}{|@{\hspace{1ex}}c@{\hspace{1ex}}||@{\hspace{1ex}}c@{\hspace{1ex}}|@{\hspace{1ex}}c@{\hspace{1ex}}|@{\hspace{1ex}}c@{\hspace{1ex}}||@{\hspace{1ex}}c@{\hspace{1ex}}|@{\hspace{1ex}}c@{\hspace{1ex}}|@{\hspace{1ex}}c@{\hspace{1ex}}||@{\hspace{1ex}}c@{\hspace{1ex}}|}
	\hline
\emph{Actual} &  \emph{Total}& \emph{Correctly} & \emph{Robustness$_{\epsilon}$}& \emph{Accuracy$_{\epsilon}$}& \emph{Accuracy$_{\epsilon}$} & \emph{Accuracy$_{\epsilon}$}& \emph{Accuracy}\\ 

\emph{Label} &  \emph{count$_{\epsilon}$}& \emph{classified count$_{\epsilon}$} & \emph{\%}& \emph{100}& \emph{1000} & \emph{10000}& \emph{(TestSet)}\\ \hline
0&\num{3.32e+270}&\num{2.21e+270}&66.67&65.00&67.90&67.05&92.65\\
1&\num{9.59e+247}&\num{3.15e+247}&32.81&34.00&32.10&32.66&96.12\\
2&\num{3.53e+258}&\num{8.81e+257}&25.00&32.00&25.90&25.30&83.91\\
3&\num{1.92e+272}&\num{1.92e+272}&99.99&93.00&93.70&93.62&77.52\\
4&\num{3.42e+264}&\num{3.42e+264}&99.99&52.00&61.50&63.58&82.08\\
5&\num{4.02e+259}&\num{4.02e+259}&99.99&100.00&100.00&100.00&76.23\\
6&\num{1.04e+258}&\num{5.22e+257}&50.00&47.40&47.40&50.21&85.39\\
7&\num{1.17e+262}&\num{1.17e+262}&99.99&100.00&100.00&100.00&85.60\\
8&\num{9.99e+266}&\num{9.99e+266}&99.99&100.00&100.00&100.00&73.72\\
9&\num{1.84e+253}&\num{6.89e+252}&37.50&38.20&38.20&38.12&80.67\\
\hline
\end{tabular}
}
\vspace{-6mm}
\end{table}

\noindent{\bf Quantifying adversarial robustness for image classification models:} We trained a decision-tree classifier on the popular MNIST benchmark, which is a collection of handwritten digits classified to one of 10 labels (0 through 9). The overall accuracy of this model on the test set was 83.64\%. We selected (randomly) an image for each of the 10 labels and considered regions around these inputs for $\epsilon = 1$; these represent all the inputs that can be generated by altering each pixel of the given image by +/- 1. Table~\ref{tab:4.3scores} presents the results. Column \emph{Total count$_\epsilon$}  shows the number of images in the $\epsilon = 1$ neighborhood of each input. We then employ \emph{QuantifyML} to  quantify the number of inputs within the $\epsilon = 1$ neighborhood that are given the correct label (Column \emph{Correctly classified count$_\epsilon$}). The corresponding \emph{Robustness$_\epsilon$} value shows the accuracy with which the model classifies the inputs in the region to the same label. The results indicate that the robustness of the model is poor or the model is more vulnerable to attacks around the inputs corresponding to labels 1, 2 and 9 respectively. 

We also computed an accuracy metric statistically by perturbing each image within $\epsilon$ to randomly generate sample sets of size 100, 1000 and 10000 images respectively. We then executed the model on each set to determine the corresponding labels and computed the respective accuracies as shown in column \emph{Accuracy}$_\epsilon$(size).  The statistically computed accuracies are close to the \emph{Robustness$_\epsilon$} values for most of the labels. However, for labels 5,7 and 8, they are 100\% respectively which gives a false impression of adversarial robustness around these inputs. The corresponding \emph{Robustness$_\epsilon$}  of 99.99\% indicates that there are subtle adversarial inputs which get missed when the robustness is determined statistically. The last column, \emph{Accuracy (TestSet)}, shows the accuracy of the model per label when evaluated statistically on the whole MNIST test set. We can observe that although the model may have high statistical accuracy, it can have low adversarial robustness. 

\noindent{\bf Quantifying the safety of machine learning classification models:}
ACAS Xu is a safety-critical collision avoidance system for unmanned aircraft control~\cite{9081758}. It receives sensor information regarding the drone (the \emph{ownship}) and any nearby intruder drones, and then issues horizontal turning advisories (one of the five labels; Clear-of-Conflict (COC), weak right, strong right, weak left, and strong left) aimed at preventing collisions. 
Previous work~\cite{Reluplex} presents 10 input-output properties that the networks need to satisfy. We used a data-set comprising of 324193 inputs and used one of the original ACAS Xu networks to obtain the labels for them. We used this dataset to train a smaller neural network with 4 layers that is amenable to a quantitative analysis. The overall accuracy of this model on the test set was 96.0\%. We selected 9 properties of ACAS Xu (see \cite{PROJGITHUB} for details on the properties) and employed our tool to evaluate the extent to which the smaller neural network model complies to each of them. 

\begin{table}[!ht]
\centering

\caption{Quantifying the safety of Neural Networks on ACAS Xu dataset. “-" shows a timeout of 5000 seconds (\emph{ApproxMC}). Properties 1 - 9 represent properties $\phi_2$ to $\phi_{10}$ from \cite{Reluplex}.}
\vspace{-3mm}
\label{tab:4.4scoresapproxmc}
\footnotesize{
\resizebox{\textwidth}{!}{
\begin{tabular}{|@{\hspace{1ex}}c@{\hspace{1ex}}||@{\hspace{1ex}}c@{\hspace{1ex}}|@{\hspace{1ex}}c@{\hspace{1ex}}|@{\hspace{1ex}}c@{\hspace{1ex}}||@{\hspace{1ex}}c@{\hspace{1ex}}|@{\hspace{1ex}}c@{\hspace{1ex}}|@{\hspace{1ex}}c@{\hspace{1ex}}||@{\hspace{1ex}}c@{\hspace{1ex}}|}
	\hline

\emph{Property} & \emph{Stat$_N$}& \emph{Stat$_S$}& \emph{Stat$_{Acc}$(\%)} & \emph{QuantifyML$_N$}& \emph{QuantifyML$_S$}& \emph{QuantifyML$_{Acc}$(\%)} & \emph{QuantifyML$_{Time}$ (s)} 

\\ \hline 

1&0&228&100.00&\num{9.00e+93}&\num{2.50e+94}&73.56&3347.1\\
2&0&0&N/A&\num{5.67e+88}&\num{2.79e+88}&32.94&4067.5\\
3&0&0&N/A&\num{3.37e+67}&\num{1.32e+65}&0.39&2791.8\\
4&0&0&N/A&\num{1.18e+74}&0&0.00&2918.4\\
5&1&4062&99.98&0&\num{2.25e+86}&100.00&1005.2\\
6&5680&140563&96.12&-&\num{6.67e+94}&-&-\\
7&0&218&100.00&0&\num{8.15e+90}&100.00&1753.5\\
8&0&1&100.00&\num{4.24e+73}&\num{8.62e+74}&95.31&2073.3\\
9&0&62&100.00&0&\num{4.17e+79}&100.00&812.2\\
\hline

\end{tabular}
}
\vspace{-4mm}
}

\end{table}

Table~\ref{tab:4.4scoresapproxmc} documents the results. We first evaluated each property statistically on a test set of size 162096 inputs (randomly selected). 
Column \emph{Stat$_N$} shows the subset of inputs in $InpSet_{P\#}$ that violate the property, \emph{Stat$_S$} shows the number of inputs in $InpSet_{P\#}$ that satisfies it, and \emph{Stat$_{Acc}$} shows the respective statistical accuracy. For each property, we calculate the \emph{QuantifyML} metrics as described in section~\ref{subsec:quantifysafety}. The \emph{QuantifyML} counts represent the portion of the input space defined by the property for which the property is satisfied or violated. For properties 2, 3 and 4, there were no inputs in the test set that belonged to the input region as defined in the property, therefore the statistical accuracy could not be calculated, whereas we were able to use \emph{QuantifyML} to evaluate the model on these properties. Results show that the neural network never satisfies property 4. This highlights the benefit of using our technique to obtain precise counts without being dependent on a set of inputs. 


\noindent{\bf On-Going work and challenges:}
\emph{MCML}~\cite{10.1145/3385412.3386015} is a tool that shares the same goal as \emph{QuantifyML} of quantification of learnability but has a dedicated implementation to decision-trees. We performed a comparison of the two tools for decision-tree models used for learning the relational graph properties. Please refer \cite{PROJGITHUB} for results. We observed that the results from the two tools matched exactly for all properties, however,  \emph{QuantifyML} is less efficient than \emph{MCML}. With projMC as the model counter, \emph{QuantifyML} takes more time for each property and times out (after 5000 secs) for three additional properties as compared to \emph{MCML}. This is because the CNF formulas generated by the CBMC tool after the analysis of the C program representation of the machine learning model is larger than that produced by \emph{MCML}, which has a custom implementation for decision-trees. We alleviated this issue by using the ApproxMC model counter, which is faster but produces approximate counts.
The analysis times for \emph{QuantifyML} are greatly reduced and we are able to obtain results for all the properties. 

The analysis of neural network models was particularly challenging. The model counters (both exact and approximate) timed out while analyzing the networks for the graph problem with more than 4 nodes. For image classification, \emph{QuantifyML} could not handle neural network models, while we could only handle a small model for ACAS Xu. For the MNIST network, we attempted to reduce the state space of the model by changing the representation of weights and biases (e.g., from \emph{floats}  to \emph{longs}). We also attempted partial evaluation by making a portion of the image pixels concrete or fixed to certain values and propagating these values to simplify computations in C program representation of the neural network. Making 10\% of the pixels concrete, led to a 51.37\% decrease in the number of variables and a 53.08\% decrease in number of clauses. However, the model counters could still not process the resulting formula in reasonable amount of time. To address the scalability problem we plan to investigate slicing and/or compositional analysis of the C program representation of the models. 

\section{Conclusion}
We presented \emph{QuantifyML} for assessing the \emph{learnability}, \emph{safety}  and \emph{robustness} of machine learning models. Our experiments show the benefit of precise quantification over statistical measures and also highlight how \emph{QuantifyML} enables comparison of different learning algorithms. 
\clearpage
\balance
\bibliographystyle{eptcs}
\bibliography{all}
\end{document}